\documentclass[runningheads]{llncs}
\usepackage[T1]{fontenc}
\usepackage{CJK}
\usepackage{amsmath}
\usepackage{graphicx}
\usepackage{multirow}
\usepackage{makecell}
\usepackage{caption}
\usepackage{subfigure}
\usepackage{paralist}
\usepackage{bbding}
\usepackage[marginal]{footmisc}

\usepackage{booktabs}
\usepackage{array}
\usepackage{pgfplotstable}
\usepackage{pgfplots} 
\usepackage{scalefnt} 

\usepackage{pgfplots} 

\begin{document}
\title{
    Dynamic Multi-View Fusion Mechanism For Chinese Relation Extraction
}
%
\author{    Jing Yang  
            \and Bin Ji \and Shasha Li
            \and Jun Ma \and Long Peng\and Jie Yu\inst{ (}\Envelope\inst{)}
}

\authorrunning{F. Author et al.}
%
\institute{
    \institute{National University of Defense Technology, Changsha, China \\
    \email{yangjing3026@alumni.nudt.edu}\\
    \email{\{jibin, shashali, majun, penglong, yj\}@nudt.edu.cn}}
}
\maketitle              
\footnote{\inst{ (}\Envelope\inst{)} Corresponding author: Jie Yu}

\begin{abstract}
%
Recently, many studies incorporate external knowledge  into character-level feature based models to improve the performance of Chinese relation extraction.
However, these methods tend to ignore the internal information of the Chinese character and cannot filter out the  noisy information of external knowledge.
To address these issues, 
we propose  a mixture-of-view-experts framework (MoVE) to  dynamically learn multi-view features for Chinese relation extraction.
%
With  both the internal and external knowledge of Chinese characters, 
our framework can better capture the semantic information of Chinese characters.
%
To demonstrate  the effectiveness of the proposed framework,
%
%
we conduct extensive experiments on three real-world datasets in distinct domains.
Experimental results show consistent and significant superiority and robustness of our proposed
framework.
%
Our code and dataset  will be released at: \url{https://gitee.com/tmg-nudt/multi-view-of-expert-for-chinese-relation-extraction}
\end{abstract}

\section{Introduction}

\begin{figure}[htbp]
\vspace{-0.51cm}
    \subfigure[The lexicon view feature.]{ 
    \begin{minipage}{6cm}
    \centering    
    \includegraphics[scale=0.58]{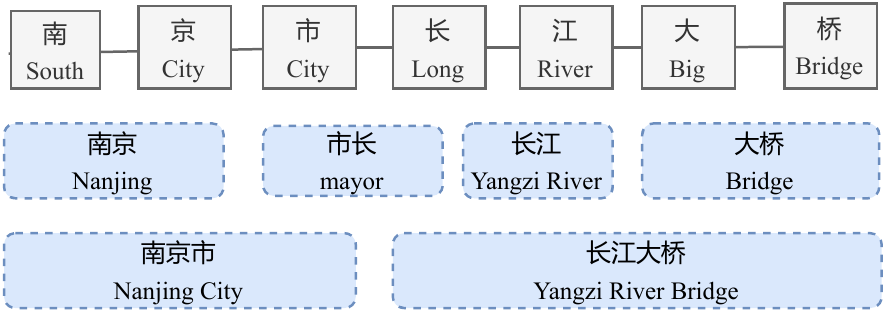}
    \label{fig:fig1a}
    \vspace{0.1cm}
    \end{minipage}
    }
    \subfigure[The radical view feature.]{   
    \begin{minipage}{6cm}
    \centering    
    \includegraphics[scale=0.58]{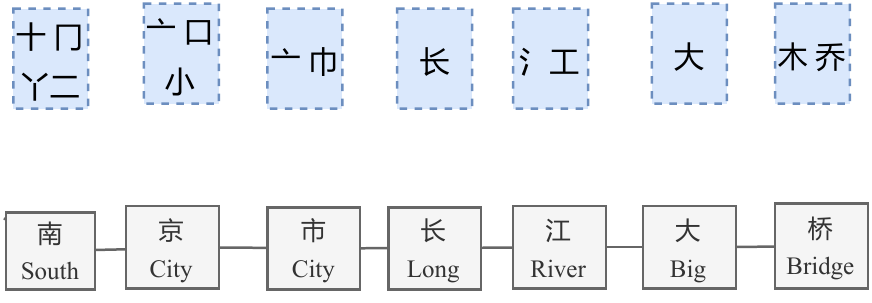}  
    \label{fig:fig1b}
    \vspace{0.2cm}
    \end{minipage}
    }

    \caption{An example of multi-view features in Chinese language texts.}   
    \label{fig: fig1}    
    \vspace{-0.8cm}
\end{figure}

Information extraction (IE) is widely considered as one of the most important topics in natural language processing (NLP), 
which is defined as  identifying the required structured information from the unstructured texts.
Relation extraction (RE) has a pivotal role in information extraction, which aims to extract semantic relations between entity pairs from unstructured texts.
%
%
%
Recently, deep learning-based models have obtained tremendous success in this task.
However, research on Chinese RE is quite limited compared to the progress in English corpora. 
We attribute  this to the following main challenge:
it is hard to extract semantic information from Chinese texts for the Chinese
language makes less use of function words and morphology.
Although there has been extensive previous work integrating the external knowledge (i.e. lexicon feature) of Chinese characters is shown to be effective for sequence labeling tasks \cite{JinxingYu2017JointEO,YuxianMeng2019GlyceGF,RuotianMa2020SimplifyTU,JintongShi2022MultilevelSF},
%
there is room for further investigation to leverage the internal characteristics of Chinese characters.
In Chinese texts, a sentence contains semantic information from different view features including character, word, structure and contextual semantic information. 
As shown in Figure \ref{fig:fig1a},  in order to reduce segmentation errors and increasing the semantic and boundary information of Chinese characters,
%
%
some methods  are proposed to establish a model to learn both character-level and word-level features \cite{RuotianMa2020SimplifyTU,ShuangWu2021MECTME}.
However, this external knowledge information is limited by the quality of domain lexicons and will inevitably introduce redundant noise.
%
%
For example, only one of \begin{CJK}{UTF8}{gbsn}{`南京}\end{CJK}(Nanjing)',\begin{CJK}{UTF8}{gbsn}{`市长}\end{CJK}(Mayor)' and  \begin{CJK}{UTF8}{gbsn}{`南京市}\end{CJK}(Nanjing City)',\begin{CJK}{UTF8}{gbsn}{`长江}\end{CJK}(YangZi River)' is an appropriate contextual information.
%
%
In addition, 
Chinese characters have evolved from pictographs since ancient times, and their structures often reflect more information about the characters.
The internal character structures  can enrich the semantic representation of Chinese characters.
As shown in Figure \ref{fig:fig1b}, \begin{CJK}{UTF8}{gbsn}{`氵'}\end{CJK}  is the radical of  \begin{CJK}{UTF8}{gbsn}{`江(River)'}\end{CJK}, and suggests \begin{CJK}{UTF8}{gbsn}{`氵(water)'}\end{CJK} that river is water-like liquid.
%
On the contrary,
the \begin{CJK}{UTF8}{gbsn}{`南(south)'}\end{CJK}  can be encoded as a structure consisting of \begin{CJK}{UTF8}{gbsn}{`十', `冂', `丫', and `二'}\end{CJK}, but they convey no meaningful semantic information.
Previous studies have proven that
semantic irrelevant sub-character component information will be noisy for representing a Chinese character \cite{JinxingYu2017JointEO,JintongShi2022MultilevelSF}.  
%
%
Although above methods have achieved reasonable performance, they still suffer from two common issue:
%
(1) The underlying fact that different view feature contains its own specific contribution to the semantic representation is ignored.
Existing methods map different view features into a shared space without interaction among views, which is difficult to guarantee that all common semantic information is adequately exploited.
(2) Existing methods introduce external knowledge as well as more noisy information, 
and they suffer from the inability of discriminating the importance of the different features and filtering out the noisy information.
%

%
%
%
Is there any other better way  to fuse multi-view features of Chinese characters? 
Inspired by mixture-of-experts \cite{NoamShazeer2017OutrageouslyLN,JiaqiMa2018ModelingTR,ZihanLiu2020ZeroResourceCN},
we propose a novel Mixture-of-View-Experts (MoVE) model to dynamically fuse both internal and external features for Chinese RE. 
As shown in Figure \ref{fig:fig2},
%
%
MoVE is a method for conditionally computing feature representation, given multiple view expert inputs that can be represented utilizing diverse knowledge sources.
%
%
In addition, a gating network is designed to dynamically calculate each expert weight per instance based on the multi-view
feature.
In this way, the knowledge from different view experts can be incorporated to model the inherent ambiguity and enhance the ability to generalize to specific domains.
%
%
%
Extensive experiments are conducted on three representative datasets across different domains,
%
%
Experimental results show that our framework consistently improves the selected baselines

In this paper, we propose a novel multi-view features fusion framework which leverages both external and internal  knowledge.
The main contributions of this paper can be summarized as follows:
\begin{compactitem}
    \item We design a novel architecture framework capable of acquiring semantic, lexical, and radical feature information from Chinese characters.
    \item Based on the multi-view features, we propose the MoVE method for dynamically composing the different features for Chinese relation extraction.
    \item Our method achieves new state-of-the-art performance on three real-world Chinese relation extraction datasets.
\end{compactitem}

\section{Related Work}

\subsection{Chinese Relation Extraction}
As a fundamental task in NLP, Relation Extraction (RE)  has been studied extensively in the past decade.
Here various neural network based models, such as  CNNs \cite{DaojianZeng2014RelationCV}, RNNs \cite{DongxuZhang2015RelationCV} or Transformer-based architectures \cite{ShanchanWu2019EnrichingPL} have been investigated.
%
Existing methods for Chinese RE are mostly character-based or word-based implementations of mainstream NRE models. 
In most cases, these methods only focus on the improvement of the model itself, 
ignoring the fact that different granularity of input will have a significant impact on the RE models. \cite{ZiranLi2019ChineseRE,JingjingXu2017ADN,QianqianZhang2018AnEG}.
The character-based model can not utilize the information of words, capturing fewer features than the word-based model. 
On the other side, the performance of the word-based model is significantly impacted by the quality of segmentation \cite{YueZhang2018ChineseNU}. 
Then, lexicon enhanced methods are used to combine character-level and word-level information in other NLP tasks like character-bigrams 
and lexicons information \cite{XiaotangZhou2022DynamicMF,HengDaXu2021ReadLA,BaojunWang2021DyLexID}.
Although, lexicon enhanced models can exploit char and external lexicon information, it still could be severely affected by the ambiguity of polysemy.
Therefore,
We utilize external linguistic knowledge  with the help of HowNet \cite{ZhendongDong2003HowNetA}, which is a concept knowledge base that annotates Chinese with correlative word synonyms.
%

\subsection{Chinese Character Representation}
Existing  models of Chinese character representation can be divided into two categories: 
exploiting the structural information of the characters themselves and injecting external knowledge.
%
%
JWE \cite{JinxingYu2017JointEO} is introduced to jointly learn Chinese component, character and word embeddings, which takes character information for improving the quality of word embeddings. 
LSN \cite{YanSong2018JointLE} is proposed to capture the relations among radicals, characters and words of Chinese and learn their embeddings synchronously.
CW2VEC \cite{CaoShaosheng2018cw2vecLC} adopts the stroke n-gram of Chinese words and utilizes the fine-grained information associated with word semantics to learn Chinese word embeddings.
In order to effectively leverage the external word semantic and enhance character boundary representation, a few models aimed to integrate lexicon information into character-level sequence labeling \cite{RuotianMa2020SimplifyTU,YueZhang2018ChineseNU,BaojunWang2021DyLexID,CanwenXu2019ExploitingME}.
Besides, there are models  utilize sense-level information with external sememe-based lexical knowledge base
,to handle the polysemy of words with the change of language situation \cite{ZiranLi2019ChineseRE,ZhendongDong2003HowNetA,FanchaoQi2019OpenHowNetAO}.

\subsection{Multi-View Learning}

There has been some research to integrate information from different multi-view to achieve better performance. 
%
ME-CNER \cite{CanwenXu2019ExploitingME} concatenates the character embeddings in radical, character and word levels to form the final character representation, which exploits multiple embeddings together in different granularities for Chinese NER.  
To fully explore the contribution of each view embedding,  FGAT \cite{XiaosuWang2021ImprovingCC} is proposed to discriminate the importance of the different granularities internal semantic features with the help of graph attention network.
ReaLiSe \cite{HengDaXu2021ReadLA}  leverages the semantic, phonetic and graphic information to tackle Chinese Spell Checking (CSC) task,  which introduce the selective fusion mechanism base Transformer \cite{AshishVaswani2017AttentionIA} to integrate multi-view information.
Recent, some efforts incorporate both internal and external multi-view information (such as lattice, glyce, pinyin, n-gram information ) with the character token in Chinese Pre-trained language models (PLMs) and design specific pre-train task \cite{ZijunSun2021ChineseBERTCP,QianglongChen2022DictBERTDD,YuxuanLai2021LatticeBERTLM}.
To the best of our knowledge, this paper is the first work to leverage multi-view information to tackle the Chinese Relation Extraction task.


\section{Methodology}

\begin{figure}
    \includegraphics[scale=0.68]{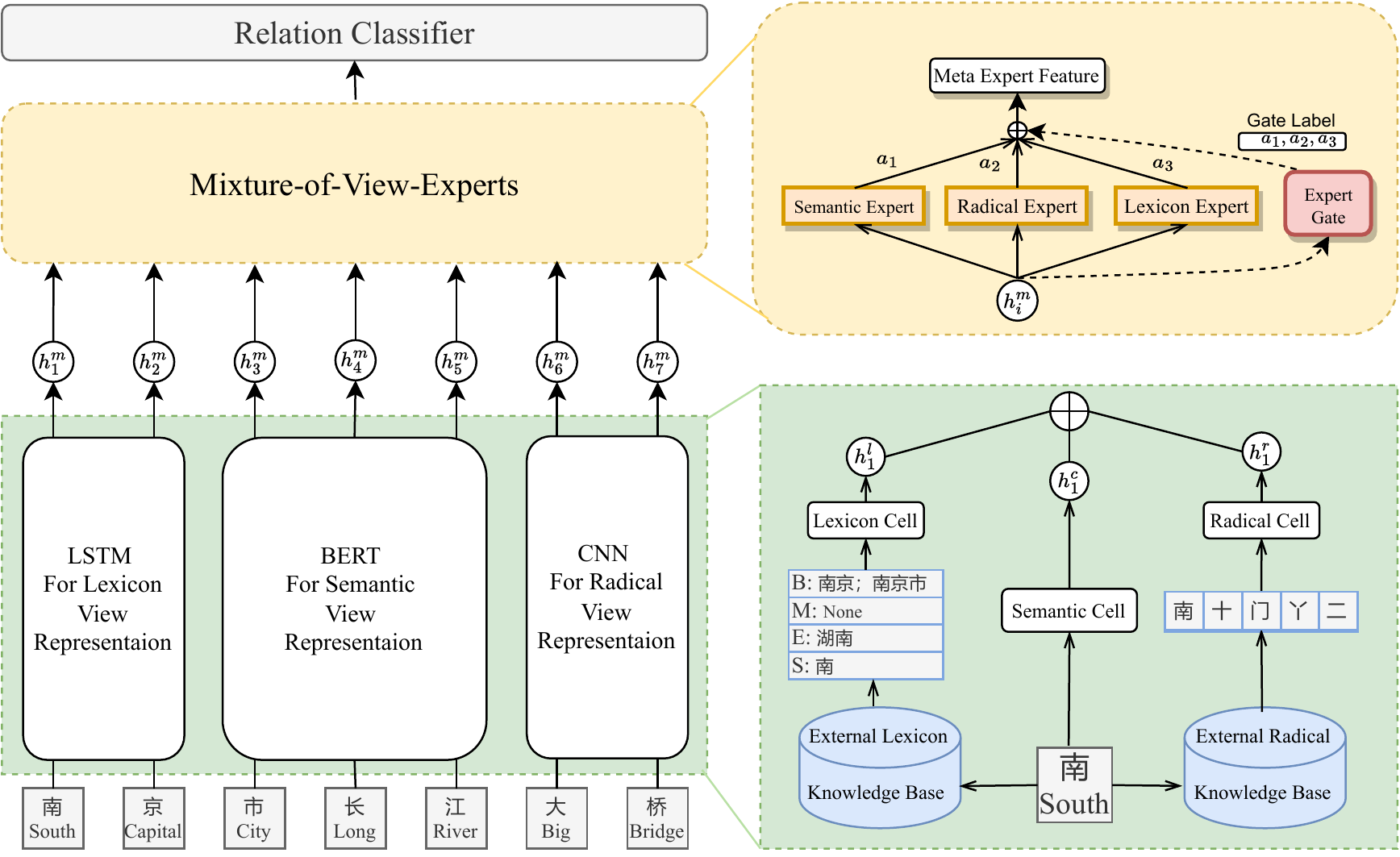}
    \caption{The architecture of  our MoVE framework. }
    \label{fig:fig2}
    \vspace{-0.3cm}
\end{figure}

%
An overview of the proposed MoVE framework is depicted in Figure \ref{fig:fig2}.
In this section,
%
we introduce our model architecture from three perspectives: Multi-View Features Representation, Mixture-of-View-Expert, 
and Relation Classifier.
%
%



\subsection{Multi-View Features Representation}
\subsubsection{Internal View Feature \newline} 
\hspace{5pt} \textbf{Semantic Embeddings}
We adopt BERT  \cite{JacobDevlin2022BERTPO} as the backbone of the semantic encoder. 
BERT provides rich contextual word representation with the unsupervised pretraining on large corpora and has been proven superior in building contextualized representations for various NLP tasks \cite{ZiranLi2019ChineseRE,HengDaXu2021ReadLA,JacobDevlin2022BERTPO,TongfengGuan2020CMeIECA}. 
%
Hence, we utilize  BERT as the underlying encoder to yield the basic contextualized character representations.
The output of the last layer  $H^c_{i}$  is used as the contextualized semantic embeddings of Chinese characters at the semantic view.
\begin{equation*}
    H^c_i = \mbox{BERT} (x_1, x_2, x_3, \dots, x_n)
\end{equation*}
%

\textbf{Radical Embeddings}  
Chinese characters are based on pictographs, and their meanings are expressed in the shape of objects.
The radical is often the semantic component and inherently bring with certain levels of semantics regardless of the contexts.
In this case, the internal structure of Chinese characters has certain useful information.
For example, the radicals such as \begin{CJK}{UTF8}{gbsn}{`月'}\end{CJK} (body) represents human body parts or organs, and \begin{CJK}{UTF8}{gbsn}{`疒'}\end{CJK}  (disease) represents diseases, which benefits Chinese  RE for the medical field.
We choose the  informative Structural Components (SC)  as radical-level features of Chinese characters, which comes from the online XinHua Dictionary\footnote{\url{https://github.com/kfcd/chaizi}}. 
Specifically, we first disassemble the Chinese characters  into $SC=(x^{c_1}_1,x^{c_2}_1, x^{c_3}_1, \dots, x^{c_i}_n )$, and then input the radical features into CNN. 
For example, we can decompose  \begin{CJK}{UTF8}{gbsn}{`脚'}\end{CJK} as \begin{CJK}{UTF8}{gbsn}{`月土厶卩'}\end{CJK}, 
\begin{CJK}{UTF8}{gbsn}{`疼'}\end{CJK} as \begin{CJK}{UTF8}{gbsn}{ `疒夂丶丶'}\end{CJK}.
Then, we use the max-pooling and fully connection layers to get the feature embedding $H^r_i$ of Chinese characters at the radical view.
\begin{equation*}
    H^r_i = \mbox{Max-Pooling}(\mbox{CNN} (x^{c_1}_1,x^{c_2}_1, x^{c_3}_1, \dots, x^{c_i}_n ))
\end{equation*}
\vspace{-1cm}
\subsubsection{External View Feature \newline}  
\hspace{5pt} \textbf{Lexicon Embeddings}
Recently, lexical enhancement methods were proposed to enhance character-based models, which have demonstrated the benefits of integrating information from external lexicons for Chinese NER tasks.
Following SoftLexicon \cite{RuotianMa2020SimplifyTU}, we  retain the segmentation information, all matched words of each character $x_i$ is categorized into four word sets `BMES'.
After obtaining the `BMES' word sets for each character, each word set is then condensed into a fixed-dimensional vector with average-pooling method:  

\begin{equation*}
    v^s(S) = \frac{1}{|S|} \sum_{w \in S} e^w (w) 
\end{equation*}
where $S \in $ `BMES' denotes a word set and $e^w$ denotes the word embedding in external lexicon.
The final step is to combine the representations of four word sets into one fix-dimension feature, and concatenate them to  get the external feature embeddings $H^l_i$ of each character at the lexicon view.

\begin{equation*}
    H^l_i = \mbox{concat}(v^s(B); v^s(M); v^s(E); v^s(S))
\end{equation*}



\vspace{-0.3cm}
\subsection{Mixture-of-View-Experts}
\vspace{-0.15cm}
%
%
%
After using the aforementioned multi-view feature embeddings methods, we get three representation at $h^c_i$,  $h^l_i$, $h^r_i$  in semantic-level, lexicon-level  and radical-level respectively.
Then we concatenate the different view features to get the multi-feature representation $h^m_i$, as shown in Figure \ref{fig:fig2}.
%
The multi-view feature representations of Chinese characters  $h^m_i$ can  capture both internal and external  features in different semantic granularity,
but they also introduce meaningless and noisy information simultaneously.
However, existing methods usually just calculate the unweighted mean of the different view features, or sometimes set the weights as hyper-parameters and calculate the weighted mean of the features \cite{RuotianMa2020SimplifyTU,XiaosuWang2021ImprovingCC}.
%
%
Moreover, existing approaches are incapable of distinguishing the significance of the introduced feature. 
%
%
Hence, they are unable to filter out introduced potential noisy information \cite{XiaotangZhou2022DynamicMF,XiaosuWang2021ImprovingCC}.

%
%
As shown in Figure \ref{fig:fig2},
we introduce an Mixture-of-View-Experts (MoVE) framework to combine representations generated by experts to produce the final prediction.
Specifically, each different feature representation acts as an view expert, which consists of two linear layers.
The expert gate consists of a linear layer followed by a softmax layer, which generates the confidence distribution over different view experts.
Finally, the meta-expert feature incorporates features from all experts based on the confidential scores from the expert gate.
We formulate the MoVE module as follows:


\begin{align*}
    [expt^{1}_{i}, \cdots,  expt^{E}_{i}] &= [L(h^m_i), \cdots, L^{E}(h^m_i)] \\
    [\alpha_{1}, \cdots, \alpha_{E}] &= \mbox{softmax}(\mbox{Linear}(h^m_i)) \\
    h^f_i &= \sum_{k=1}^{E} \alpha_{k} * expt^{k}_i 
\end{align*}    
where $h^f_i$ is the meta-expert feature of $h_i$,  $expt$ is the feature derived from the expert, and L denotes the linear layer. 
As shown in Figure \ref{fig:fig2},
%
%
the MoVE has three types of experts, namely semantic, radical, and lexicon experts.
The expert features are computed based on the multi-view feature representations,
and the predictions are conditioned on the meta-expert features and the multi-view features.
%
%

\subsection{Relation Classifier}
After the MoVE model dynamically combines meta-expert feature $h^f_i$ for each token.
We first merge $h^f_i$ into a sentence-level feature vector $H^f$,
then the final sentence representation $H^f$ is  passed through  a softmax classifier to compute the confidence of each relation.
We formulate the classifier module as follows:

\begin{equation*}
    P(y|S) = \mbox{softmax}(W \cdot H^f + b)
\end{equation*}
where $W \in R^{Y \times d} $ is the transformation matrix and $b \in R^Y$ is a bias vector. $Y$ indicates the total number of relation types, and $y$ is the estimated probability for each type. 
%
%
%
Finally, given all training examples $(S^i, y^i)$, we define the objective
function using the following cross-entropy loss:

\begin{equation*}
    \mathcal{L} (\theta) = \sum_{i=1}^{T} log P(y^i | S^i, \theta)
\end{equation*}
\vspace{-1cm}

\vspace{-0.3cm}
\section{Experiments}
\vspace{-0.15cm}
\subsection{Datasets}
We evaluate our approach on two popular Chinese RE datasets: FinRE \cite{ZiranLi2019ChineseRE} and  SanWen \cite{JingjingXu2017ADN}.
To increase domain diversity, we manually annotate the SciRE, which is the first Chinese  dataset for scientific relation extraction.
%
%
FinRE  is a manual-labeled financial news dataset, which contains 44 distinguished relationships, including a special relation NA.
SanWen is a document-level Chinese literature dataset for relation extraction, including  9 relation types specific to Chinese literature articles.
%
%
The SciRE dataset is collected from  3500 Chinese scientific papers in CNKI\footnote{\url{https://www.cnki.net/}},
which defines scientific terms and relations especially for scientific knowledge graph construction.
There are  4 relation types (Used-For, Compare-For, Conjunction-Of, Hyponym-Of)  defined in SciRE dataset.
%
%
Following previous work, we use the same preprocessing procedure and splits for all datasets \cite{ZiranLi2019ChineseRE}.
Table \ref{tab: tab1} shows the characteristics of each dataset.

\begin{table}
\centering
\vspace{-0.8cm}
\caption{Statistics of the three experimental datasets.}
\vspace{0.2cm}
\label{tab: tab1}
\begin{tabular}{p{1.5cm}|p{1.2cm}|p{1.8cm}|p{2.2cm}|p{1.2cm}|p{1.2cm}|p{1.2cm}}
\toprule
    Dataset  &$\#$Type &Domain  &Characteristic  &\#Train  &\#Dev  &\#Test  \\
    \midrule
        \multirow{2}{*}{ \makecell{FinRE }}   
        &\multirow{2}{*}{ \makecell{44}} &\multirow{2}{*}{ \makecell{Financial}} &Sentences   &4477   &500    &1219  \\
        &                                &                                        &Triples   &9873   &1105   &2722  \\
        
    \hline 
    \specialrule{0em}{1pt}{1pt}
        \multirow{2}{*}{ \makecell{SanWen }}   
        &\multirow{2}{*}{ \makecell{9}} &\multirow{2}{*}{ \makecell{Literature}} &Sentences   &10754   &1108    &1376\\
        &                                &                                        &Triples   &12608   &1283    &1560\\
        
    \hline
    \specialrule{0em}{1pt}{1pt}
        \multirow{2}{*}{ \makecell{SciRE }}   
        &\multirow{2}{*}{ \makecell{4}} &\multirow{2}{*}{ \makecell{Scientific}}  &Sentences  &7251    &1067    &1990\\
        &                                &                                        &Triples   &18548   &2778    &5148\\
 
    \bottomrule
\end{tabular}
\vspace{-0.8cm}
\end{table}

\subsection{Baselines}
To investigate the effectiveness of our model, we compare our model with the both character-based and lattice-based variations of the five following models.
For the character-based models, we conduct  experiments with BLSTM \cite{DongxuZhang2015RelationCV}, Att-BLSTM \cite{PengZhou2016AttentionBasedBL}, PCNN \cite{YankaiLin2016NeuralRE} and Att-PCNN \cite{JoohongLee2019SemanticRC},  which utilize traditional neural network  RNN, CNN or Attention mechanism for Chinese relation extraction.
%
%
We use DeepKE \cite{NingyuZhang2022DeepKEAD}, an open-source neural network relation extraction toolkit  to conduct the experiments.
For the lattice-based models, we compare with Basic-Lattice and MG-Lattice. 
In Basic-Lattice \cite{YueZhang2018ChineseNU}, an extra word cell is employed to encode the potential words, and attention mechanism is used to fuse feature,  which can explicitly leverages character and word information. 
Moreover, MG-Lattice \cite{ZiranLi2019ChineseRE} models multiple senses of polysemous words with the help of external linguistic knowledge 
to alleviate polysemy ambiguity .
%

%
%
We use Chines BERT-wwm \cite{YimingCui2021PreTrainingWW} as the base semantic encoder for all datasets.
We follow the standard evaluation metric and report Precision, Recall, and F1 scores to compare the performance of different models.
For each view feature, we implement a linear projection layer with  hidden dim 100.
We train our model with the AdamW \cite{IlyaLoshchilov2018FixingWD} optimizer for 50 epochs. 
The learning rate is set to 1e-3/5e-5, the batch size is set to 32, 
and the model is trained with learning rate warmup ratio 10\% and linear decay.
%
%
\subsection{Experimental Results}
To conduct a comprehensive comparison and analysis, we conduct experiments on character-based, lattice-based, multiview-based models on three datasets.
%
%
For a fair comparison,  we implement Basic-MultiView semantic encoder  by replacing the BERT with a bidirectional LSTM and improving the representation of characters using additional bi-word features.
In addition, to verify the capability of our method combined with the pretrained model, we choose our method  with the BERT+BLSTM model.

\begin{table}
\centering
\vspace{-0.8cm}
\caption{F1-scores of Character, Lattice and Multi-View models on all datasets.}
\vspace{0.1cm}
\label{tab: tab2}
\begin{tabular}{p{4cm}|p{3cm}|p{1.5cm}|p{1.5cm}|p{1.5cm}}
\toprule
    \multicolumn{2}{c|}{Baseline} &  FinRE & SanWen & SciRE \\
    


    \midrule    
        
    \multirow{4}{*}{ \makecell{Character-based }}   
        &BLSTM       &42.87   &61.04   &87.35\\
        &Att-BLSTM  &41.48   &59.48   &88.47\\
        &PCNN        &45.51   &81.00  &87.86\\
        &Att-PCNN    &46.13   &60.55  &88.78\\

    
    \hline
    \specialrule{0em}{1pt}{1pt}
    \multirow{2}{*}{ \makecell{Lattice-based }}   
        &Basic-Lattice  &47.41   &63.88  &89.25\\
        &MG-Lattice     &49.26   &65.61  &89.82\\

    \hline 
    \specialrule{0em}{1pt}{1pt}
    \multirow{2}{*}{ \makecell{MultiView-based }}  
        &Basic-MultiView     &51.01   &\textbf{68.96}     &90.32  \\
        &MultiView+biword    &\textbf{51.56}    &68.45     &\textbf{91.32}\\
    
    \hline 
    \specialrule{0em}{1pt}{1pt}
    \multirow{2}{*}{ \makecell{MultiView-(BERT)}}  
        &BERT+BLSTM        &51.43    &70.12  &91.25\\
        &BERT+MultiView    &\textbf{53.89}    &\textbf{72.98}  &\textbf{92.18}\\

    \bottomrule
\end{tabular}
\vspace{-0.5cm}
\end{table}



%
%
We report the main results in Table \ref{tab: tab2}, from which we can observe that:
%
%
(1)
The performance of the MultiView-based  model that integrates radical feature and lexicon feature into character-based RE models is notably better than the previous RE baseline models. 
This shows our model can leverage the  knowledge from character, radical, and lexicon effectively.
%
(2)
We observe that our model consistently outperforms the  character-based and lattice-based model on three datasets.
%
(3)
We can see that the BERT-MultiView outperforms the BERT-BLSTM, which show that the our method can be effectively combined with pre-trained model. 
Moreover, the results also verify the effectiveness of our method in utilizing multi-view features  information, which means it can complement the information obtained from the pre-trained model.
Based on the experimental results above,
it makes sense that integrating different granularity and view information and pre-trained model is beneficial for Chinese RE. 

\subsection{Ablation Studies}
We conduct ablation studies to further investigate the effectiveness of the main components in our model on SanWen from two perspectives:  Multi-View Features Encoder Layer and Mixture-of-View-Experts (MoVE) fusion mechanism.

\textbf{Effect against  Multi-View Features Encoder Layer}.
In this part, we mainly focus on the effect of the different view encoder layer and we conduct following experiment with Basic-MultiView. 
We consider three model variants setting:  w/o semantic layer, w/o lexicon layer, w/o radical layer.
%


\begin{table}
\centering
\vspace{-0.8cm}
\caption{Precision, Recall, and F1 score on SanWen.}
\vspace{0.1cm}
\label{tab: tab3}
\begin{tabular}{p{5cm}|p{2cm}|p{2cm}|p{2cm}}
    \toprule
         Model  Variant &Precision &Recall  &F1  \\
        \midrule
            Basic-MultiView   &68.55  &67.89  &68.96   \\
        \hline
        \specialrule{0em}{1pt}{1pt}
            \quad w/o Semantic View   &49.47   &53.69  &51.50 \\
            \quad w/o Lexicon View    &64.29    &67.88  &66.04 \\
            \quad w/o Radical View    &66.78    &66.58  &66.68 \\
        \bottomrule
    \end{tabular}
\vspace{-0.5cm}
\end{table}
%

As shown in Table \ref{tab: tab3},
the performance of our model degrades regardless of which view encoder  removed.
The model without semantic layer has the most significant performance drops,  which means that contextual information  between entities is the most important feature in our multi-view encoder.
The lexicon features are complementary to  character-based semantic information in different granularity,
which also is consistent with previous studies \cite{RuotianMa2020SimplifyTU,CanwenXu2019ExploitingME}.
Finally, we are surprised  to find that the less-mentioned radical features, compared to the lexicon feature, appear to bring more benefits.
We conclude  that radical information that will give more concrete evidence and show more logic patterns to  the model.
%

\begin{figure}
    \includegraphics[scale=0.245]{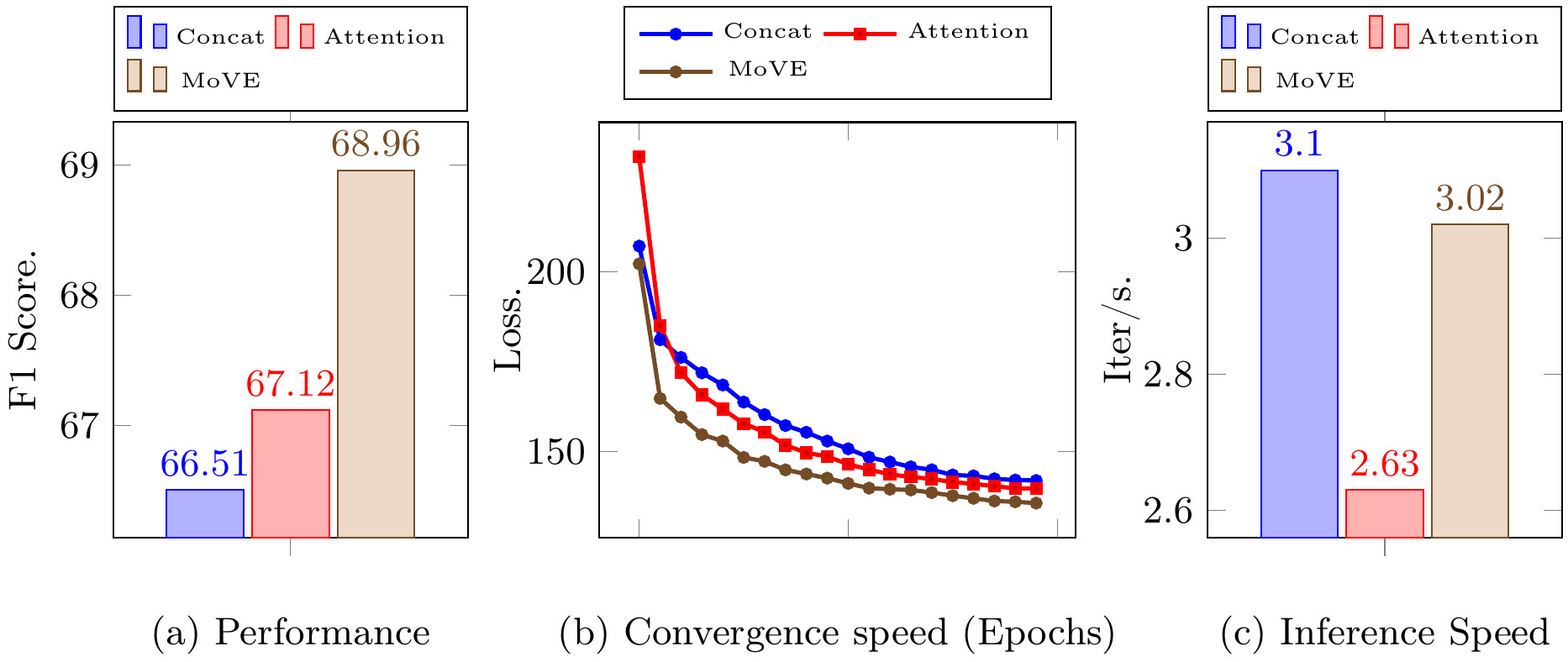}
    \caption{Compared with different view fusion strategies in Performance, Conver-
gence Speed, Inference Speed. }
    \label{fig:fig3}
    \vspace{-0.3cm}
\end{figure}

%
\textbf{Effect against MoVE Fusion Layer.} 
In this part, we compare three different multi-view fusion strategies: concat (Concat), attention (Attention) and our mixture-of-view-experts (MoVE).
%
The most intuitive way is Concat method, which just put the different view feature together and feed to classifier layer.
In order to interact with the information of different views, the existing work  design a fusion layer based on the Attention mechanism \cite{AshishVaswani2017AttentionIA}.
As shown in Figure \ref{fig:fig3}(a), we find that the performance of our designed fusion module exceeds Concat 1.2\% and Attention 0.5\% on SanWen, respectively.
%
%
This is due to the fact that the proposed MoVE can adapt the fusion weight of relevant views based on the properties of datasets, allowing the model to get appropriate composite features.
%

    

A potential concern of our model is that the implementation of MoVE brings additional parameters and increases the complexity of the model.
To verify the impact of the MoVE module on the efficiency of our model, 
we use the same  hyper-parameters  setting to observe the convergence speed of the model during training and the inference speed during prediction under three different view fusion strategies.
As show in Figure \ref{fig:fig3}(b), 
with the same training settings, the MoVE helps the model to converge faster and be more stable than Concat and Attention strategies.
We attribute this to the ability of MoVE to fuse different view information and filter noise information more effectively, which helps the model to learn more efficiently.
Figure \ref{fig:fig3}(c) shows the inference speed.
Our MoVE does not reduce the inference speed, because in inference stage the gating structure of MoVE only sparsely activated,
and will only leverage information from views that have made significant contributions in the training phase.
%

%


\section{Conclusion}
In this paper, we propose a novel multi-view features model for Chinese relation extraction.
The proposed model integrates different view representations that fuses semantic,  lexicon  and radical features through a mixture-of-view-expert mechanism.
%
%
We conduct extensive experiments on three distinct domain datasets and compare our model with several strong baselines. 
Experimental results demonstrate that the multi-view Chinese characters can effectively improve the performance for Chinese RE.
Furthermore, the ablation studies show  mixture-of-view-expert mechanism  effectively  filters noisy information while improving the efficiency of model training and inference.
%
In further, 
we will consider how to integrate  more knowledge such as part-of-speech, dependency and syntax information and extend it to other language and NLP tasks.

\section*{Acknowledgments.}
This research was supported partly by the science and technology innovation program of Hunan province under grant No. 2021GK2001.

%
%
%
\bibliographystyle{splncs04_unsort}
\bibliography{reference.bib}

\begin{thebibliography}{10}
\providecommand{\url}[1]{\texttt{#1}}
\providecommand{\urlprefix}{URL }
\providecommand{\doi}[1]{https://doi.org/#1}

\bibitem{JinxingYu2017JointEO}
Yu, J., Jian, X., Xin, H., Song, Y.: Joint embeddings of chinese words,
  characters, and fine-grained subcharacter components. empirical methods in
  natural language processing  (2017)

\bibitem{YuxianMeng2019GlyceGF}
Meng, Y., Wu, W., Wang, F., Li, X., Nie, P., Yin, F., Li, M., Han, Q., Sun, X.,
  Li, J.: Glyce: Glyph-vectors for chinese character representations. neural
  information processing systems  (2019)

\bibitem{RuotianMa2020SimplifyTU}
Ma, R., Peng, M., Zhang, Q., Wei, Z., Huang, X.: Simplify the usage of lexicon
  in chinese ner. meeting of the association for computational linguistics
  (2020)

\bibitem{JintongShi2022MultilevelSF}
Shi, J., Sun, M., Sun, Z., Li, M., Gu, Y., Zhang, W.: Multi-level semantic
  fusion network for chinese medical named entity recognition (2022)

\bibitem{ShuangWu2021MECTME}
Wu, S., Song, X., Feng, Z.H.: Mect: Multi-metadata embedding based
  cross-transformer for chinese named entity recognition. meeting of the
  association for computational linguistics  (2021)

\bibitem{NoamShazeer2017OutrageouslyLN}
Shazeer, N., Mirhoseini, A., Maziarz, K., Davis, A., Le, Q.V., Hinton, G.E.,
  Dean, J.: Outrageously large neural networks: The sparsely-gated
  mixture-of-experts layer. Learning  (2017)

\bibitem{JiaqiMa2018ModelingTR}
Ma, J., Zhao, Z., Yi, X., Chen, J., Hong, L., Chi, E.H.: Modeling task
  relationships in multi-task learning with multi-gate mixture-of-experts.
  knowledge discovery and data mining  (2018)

\bibitem{ZihanLiu2020ZeroResourceCN}
Liu, Z., Winata, G.I., Fung, P.: Zero-resource cross-domain named entity
  recognition. meeting of the association for computational linguistics  (2020)

\bibitem{DaojianZeng2014RelationCV}
Zeng, D., Liu, K., Lai, S., Zhou, G., Zhao, J.: Relation classification via
  convolutional deep neural network. international conference on computational
  linguistics  (2014)

\bibitem{DongxuZhang2015RelationCV}
Zhang, D., Wang, D.: Relation classification via recurrent neural network.
  arXiv: Computation and Language  (2015)

\bibitem{ShanchanWu2019EnrichingPL}
Wu, S., He, Y.: Enriching pre-trained language model with entity information
  for relation classification. conference on information and knowledge
  management  (2019)

\bibitem{ZiranLi2019ChineseRE}
Li, Z., Ding, N., Liu, Z., Zheng, H.T., Shen, Y.: Chinese relation extraction
  with multi-grained information and external linguistic knowledge. meeting of
  the association for computational linguistics  (2019)

\bibitem{JingjingXu2017ADN}
Xu, J., Wen, J., Sun, X., Su, Q.: A discourse-level named entity recognition
  and relation extraction dataset for chinese literature text. arXiv:
  Computation and Language  (2017)

\bibitem{QianqianZhang2018AnEG}
qian Zhang, Q., dong Chen, M., zhong Liu, L.: An effective gated recurrent unit
  network model for chinese relation extraction. DEStech Transactions on
  Computer Science and Engineering  (2018)

\bibitem{YueZhang2018ChineseNU}
Zhang, Y., Yang, J.: Chinese ner using lattice lstm. meeting of the association
  for computational linguistics  (2018)

\bibitem{XiaotangZhou2022DynamicMF}
Zhou, X., Zhang, T., Cheng, C., Song, S.: Dynamic multichannel fusion mechanism
  based on a graph attention network and bert for aspect-based sentiment
  classification (2022)

\bibitem{HengDaXu2021ReadLA}
Xu, H.D., Li, Z., Zhou, Q., Li, C., Wang, Z., Cao, Y., Huang, H., Mao, X.L.:
  Read, listen, and see: Leveraging multimodal information helps chinese spell
  checking. meeting of the association for computational linguistics  (2021)

\bibitem{BaojunWang2021DyLexID}
Wang, B., Zhang, Z., Xu, K., Hao, G.Y., Zhang, Y., Shang, L., Li, L., Chen, X.,
  Jiang, X., Liu, Q.: Dylex: Incorporating dynamic lexicons into bert for
  sequence labeling. empirical methods in natural language processing  (2021)

\bibitem{ZhendongDong2003HowNetA}
Dong, Z., Dong, Q.: Hownet - a hybrid language and knowledge resource.
  international conference natural language processing  (2003)

\bibitem{YanSong2018JointLE}
Song, Y., Shi, S., Li, J.: Joint learning embeddings for chinese words and
  their components via ladder structured networks. international joint
  conference on artificial intelligence  (2018)

\bibitem{CaoShaosheng2018cw2vecLC}
Shaosheng, C., Lu, W., Zhou, J., Li, X.: cw2vec: Learning chinese word
  embeddings with stroke n-gram information. national conference on artificial
  intelligence  (2018)

\bibitem{CanwenXu2019ExploitingME}
Xu, C., Wang, F., Han, J., Li, C.: Exploiting multiple embeddings for chinese
  named entity recognition. conference on information and knowledge management
  (2019)

\bibitem{FanchaoQi2019OpenHowNetAO}
Qi, F., Yang, C., Liu, Z., Dong, Q., Sun, M., Dong, Z.: Openhownet: An open
  sememe-based lexical knowledge base. arXiv: Computation and Language  (2019)

\bibitem{XiaosuWang2021ImprovingCC}
Wang, X., Xiong, Y., Niu, H., Yue, J., Zhu, Y., Yu, P.S.: Improving chinese
  character representation with formation graph attention network. conference
  on information and knowledge management  (2021)

\bibitem{AshishVaswani2017AttentionIA}
Vaswani, A., Shazeer, N., Parmar, N., Uszkoreit, J., Jones, L., Gomez, A.N.,
  Kaiser, L., Polosukhin, I.: Attention is all you need. neural information
  processing systems  (2017)

\bibitem{ZijunSun2021ChineseBERTCP}
Sun, Z., Li, X., Sun, X., Meng, Y., Ao, X., He, Q., Wu, F., Li, J.:
  Chinesebert: Chinese pretraining enhanced by glyph and pinyin information.
  meeting of the association for computational linguistics  (2021)

\bibitem{QianglongChen2022DictBERTDD}
Chen, Q., Li, F.L., Xu, G., Yan, M., Zhang, J., Zhang, Y.: Dictbert: Dictionary
  description knowledge enhanced language model pre-training via contrastive
  learning. international joint conference on artificial intelligence  (2022)

\bibitem{YuxuanLai2021LatticeBERTLM}
Lai, Y., Liu, Y., Feng, Y., Huang, S., Zhao, D.: Lattice-bert: Leveraging
  multi-granularity representations in chinese pre-trained language models.
  north american chapter of the association for computational linguistics
  (2021)

\bibitem{JacobDevlin2022BERTPO}
Devlin, J., Chang, M.W., Lee, K., Toutanova, K.: Bert: Pre-training of deep
  bidirectional transformers for language understanding (2022)

\bibitem{TongfengGuan2020CMeIECA}
Guan, T., Zan, H., Zhou, X., Xu, H., Zhang, K.: Cmeie: Construction and
  evaluation of chinese medical information extraction dataset. international
  conference natural language processing  (2020)

\bibitem{PengZhou2016AttentionBasedBL}
Zhou, P., Shi, W., Tian, J., Qi, Z., Li, B., Hongwei, H., Xu, B.:
  Attention-based bidirectional long short-term memory networks for relation
  classification. meeting of the association for computational linguistics
  (2016)

\bibitem{YankaiLin2016NeuralRE}
Lin, Y., Shen, S., Liu, Z., Luan, H., Sun, M.: Neural relation extraction with
  selective attention over instances. meeting of the association for
  computational linguistics  (2016)

\bibitem{JoohongLee2019SemanticRC}
Lee, J., Seo, S., Choi, Y.S.: Semantic relation classification via
  bidirectional lstm networks with entity-aware attention using latent entity
  typing. Symmetry  (2019)

\bibitem{NingyuZhang2022DeepKEAD}
Zhang, N., Xu, X., Tao, L., Yu, H., Ye, H., Xie, X., Chen, X., Li, Z., Li, L.,
  Liang, X., Yao, Y., Deng, S., Zhang, W., Zhang, Z., Tan, C., Huang, F.,
  Zheng, G., Chen, H.: Deepke: A deep learning based knowledge extraction
  toolkit for knowledge base population (2022)

\bibitem{YimingCui2021PreTrainingWW}
Cui, Y., Che, W., Liu, T., Qin, B., Yang, Z., Wang, S., Hu, G.: Pre-training
  with whole word masking for chinese bert. IEEE Transactions on Audio, Speech,
  and Language Processing  (2021)

\bibitem{IlyaLoshchilov2018FixingWD}
Loshchilov, I., Hutter, F.: Fixing weight decay regularization in adam (2018)

\end{thebibliography}
\end{document}